\documentclass{article}

\usepackage[preprint]{ewrl_2026}

\usepackage{comment}

\usepackage[utf8]{inputenc} 
\usepackage[T1]{fontenc}    
\usepackage{hyperref}       
\usepackage{url}            

\usepackage{amsfonts}       
\usepackage{nicefrac}       
\usepackage{microtype}      
\usepackage{xcolor}         

\usepackage{bbm}

\usepackage{graphicx}
\usepackage{subfigure}
\usepackage{xspace}

\usepackage{multirow}
\usepackage{pifont}

\usepackage{tabularx}
\usepackage{array}

\usepackage{amsmath}
\usepackage{amssymb}
\usepackage{mathtools}
\usepackage{amsthm}

\usepackage{svg}

\usepackage[capitalize,noabbrev]{cleveref}

\usepackage[textsize=tiny]{todonotes}

\usepackage{booktabs}       
\hypersetup{colorlinks=true, linkcolor=blue, citecolor=teal, urlcolor=magenta}

\newcommand{\cmark}{\textcolor{green!60!black}{\ding{51}}}
\newcommand{\xmark}{\textcolor{red}{\ding{55}}}

\newcommand{\method}{\textsc{MODIP}\xspace} 

\newcommand{\diffpol}{DP\xspace}
\newcommand{\diffpols}{DPs\xspace}

\newcommand{\Hmpc}{H_{\text{mpc}}\xspace}

\title{\method: Efficient Model-Based Optimization for Diffusion Policies}

\author{%
  Zakariae El~Asri $^{1}$
        \quad
        Philippe Gratias-Quiquandon $^{1}$  
        \quad
        Nicolas Thome $^{1,2}$
        \quad
        Olivier Sigaud $^{1}$\\
    $^{1}$ Sorbonne Universit\'e, CNRS, ISIR, F-75005 Paris, France\\
    $^{2}$  Institut Universitaire de France (IUF)\\
  \texttt{\{zakariae.el\_asri,  philippe.gratias\_quiquandon, olivier.sigaud, nicolas.thome\}}\\\texttt{@sorbonne-universite.fr} \\
  }

\begin{document}

\maketitle

\begin{abstract}
Diffusion policies (DPs) have emerged as expressive policy representations for robot learning, often used with imitation learning methods such as behavioral cloning (BC). However, while their success has largely been confined to BC, direct reinforcement learning (RL) fine-tuning remains challenging because actions are generated through a multi-step denoising process. In this work, we propose \method, a framework for the offline-to-online fine-tuning of DPs.
Rather than directly applying RL to the \diffpol, \method leverages a world model (WM) to guide policy adaptation and keeps the simplicity and stability of BC. We utilize model predictive control (MPC) to generate high-quality trajectories within the WM, and use them as supervised targets for fine-tuning the DP. To make MPC planning efficient, \method uses a terminal state value instead of a policy-dependent state-action value, reducing inference time. 
Additionally, \method trains critics with policy-independent TD targets, reducing training time. Experiments on D4RL (MuJoCo, Kitchen) and RoboMimic tasks show that \method improves diffusion policies beyond BC, and is competitive with or outperforms diffusion policy RL fine-tuning methods and strong model-based baselines such as TD-MPC2.
Code is available at: \url{github.com/elasriz/DPMPC/} 
\end{abstract}

\section{Introduction}
Demonstration datasets are becoming increasingly available for robot learning, making it possible to train policies directly from diverse offline data \citep{xiao2025robot}. Standard MLP policies, often instantiated as deterministic or Gaussian controllers, have achieved strong performance in many continuous control and robotic tasks due to their simplicity. However, these policies typically represent unimodal action distributions, which can be limiting when demonstrations contain multiple valid behaviors to achieve the same task. This is especially important in robotic manipulation, where tasks often admit multiple valid strategies corresponding to different action sequences from the same context \citep{chi2023diffusion}. Therefore, effective learning from multimodal demonstrations requires more expressive policy representations.

The diffusion policy (\diffpol) framework has recently emerged as a powerful alternative for imitation learning and robotic control \citep{chi2023diffusion}. By generating action sequences through an iterative denoising process, they can model complex and multi-modal behavior distributions while retaining a stable supervised training objective such as the behavioral cloning (BC) loss function. This makes them especially well-suited for learning from demonstration datasets in manipulation tasks. However, when trained only with imitation learning, \diffpols inherit its fundamental limitation: they may reproduce the behavior present in the dataset, but do not directly improve beyond it.

When a reward signal is available, reinforcement learning (RL) provides a natural way to improve a pretrained \diffpol beyond imitation learning. Recent RL methods such as Diffusion-QL \citep{wang2022diffusion}, EDP \citep{Kang2023EfficientDP}, DPPO \citep{ren2025diffusion}, and DSRL \citep{wagenmaker2025steering} show that \diffpols can indeed be improved using RL. However, fine-tuning \diffpols with RL remains more challenging than fine-tuning standard Gaussian policies. Since actions are generated through a multi-step denoising process, policy optimization is more complex and may suffer from instability.

In this work, we take a different approach. Rather than directly optimizing the \diffpol with RL, we use model predictive control (MPC) and BC as an indirect policy improvement operator. Our method, \method, learns a latent world model, a reward model, and a terminal value function. At interaction time, a DP-guided MPC planner uses the \diffpol as an expressive action sequence prior within a sampling-based trajectory optimizer such as Model Predictive Path Integral control (MPPI) \citep{williams2017information}. The planner combines diffusion-generated candidates with randomly sampled trajectories and refines them through model-based rollouts in the learned latent space. It scores them using predicted rewards together with the terminal value $V$. The resulting trajectories are then used to fine-tune the \diffpol with the standard supervised denoising loss. As a result, the policy retains the simplicity and stability of supervised learning, while improving through a progressively better training data distribution induced by MPC.

A key challenge in combining \diffpols with planning is computational efficiency. 
Standard hybrid planners \citep{Hansen2022tdmpc} often use a terminal estimate of the form $Q(s,\pi(s))$ to capture long-term returns exceeding the planning horizon for a given state $s$ and policy $\pi(s)$. When $\pi$ is a \diffpol, this requires running the denoising process in the terminal state, increasing the inference time. We instead use a terminal state-value function $V(s)$, which provides a single forward-pass estimate of the long-horizon return. Besides, we decouple critic learning from the current policy by forming temporal difference targets using target state values rather than actions sampled from the \diffpol. This avoids calling the \diffpol during critic updates and reduces training time.

Our contributions are as follows.
\par$\bullet$  \textbf{Hybrid DP-guided MPC planning.}
    We introduce a DP-guided MPC planner that uses a \diffpol as a multi-modal action sequence prior and refines its proposals through latent MPC.
\par$\bullet$  \textbf{MPC-to-policy distillation.}
    We fine-tune the \diffpol by BC on trajectories improved by the DP-guided MPC planner.
\par$\bullet$ \textbf{Efficient value-based trajectory scoring.}
    We replace the standard terminal estimate $Q(s,\pi(s))$ with a terminal state value $V(s)$, avoiding \diffpol queries at terminal states and reducing planning-time inference cost.
\par$\bullet$  \textbf{Policy-independent critic targets.}
    We decouple value learning from the current \diffpol by forming TD targets using state-value estimates instead of policy-sampled actions, reducing the cost of critic updates.

\section{Related Work}
Our work lies at the intersection of \diffpols, latent model-based planning, and offline-to-online policy improvement. 
\paragraph{Diffusion policies for robot learning.}
Diffusion models have recently been introduced as expressive policy representations for robotic control. \diffpol \citep{chi2023diffusion} formulates visuomotor control as conditional denoising over action sequences and shows strong performance on manipulation tasks, partly due to its ability to represent multi-modal action distributions. This expressiveness is particularly useful in imitation learning, where demonstrations may contain several valid modes of behavior. Our work builds on this policy class but focuses on improving \diffpols beyond offline imitation learning through model-based planning and distillation.

\paragraph{Reinforcement learning with diffusion policies.}
Several recent methods have explored how to combine \diffpols with RL. Diffusion-QL \citep{wang2022diffusion} uses a \diffpol in offline RL and augments the denoising objective with value maximization. Efficient Diffusion Policy \citep{Kang2023EfficientDP} reduces the computational cost of \diffpol training for offline RL by avoiding full sampling chains during training. DPPO \citep{ren2025diffusion} studies policy-gradient fine-tuning of pretrained \diffpols and shows that \diffpols can be improved with online RL. Other approaches, such as DSRL \citep{wagenmaker2025steering}, adapt \diffpols through latent space RL. In contrast to these methods, \method does not directly optimize the \diffpol with an RL objective. Instead, it uses a model-based planner to generate improved trajectories and trains the \diffpol with a standard supervised denoising loss.

Closest to our approach, PA-RL \citep{mark2024policy} improves expressive policies by first optimizing candidate actions with a critic and then distilling the optimized actions into the policy. These optimized actions are obtained by sampling from a base policy and improving candidates through critic-based optimization. \method shares with PA-RL the idea of improving expressive policies through supervised learning on improved actions, but optimizes actions through MPC-based optimization rather than a critic-based one, which performs better.

\paragraph{Latent model-based planning and hybrid MPC.}
Model-based RL (MBRL) improves sample efficiency by learning a predictive model of the environment and using it for planning or policy improvement, as in PETS \citep{chua2018deep} and MBPO \citep{janner2019trust}. In this work, we focus on latent model-based planners and hybrid MPC methods, which are more directly related to our approach. TD-MPC \citep{Hansen2022tdmpc} learns a task-oriented latent dynamics model and performs trajectory optimization directly in the latent space, using a learned terminal value function to account for long-horizon returns beyond the short planning horizon. TD-MPC2 \citep{hansen2024td} further scales this paradigm with improved architectures and training procedures, showing that latent MPC can be effective in various continuous control tasks. Existing hybrid planners commonly use policy-dependent terminal estimates such as $Q(s,\pi(s))$, which become expensive when $\pi$ is an iterative \diffpol. In contrast, BMPC \citep{wang2025bootstrapped} uses a terminal state value $V(s)$ as \method does. However, \method goes one step further with its policy-independent critic learning mechanism.

\paragraph{Offline-to-online reinforcement learning.}
Offline-to-online RL aims to leverage prior datasets to initialize agents before online fine-tuning. A key design choice in this setting is whether critic learning depends on the current policy. IQL \citep{kostrikov2022iql} follows a policy-independent approach: it learns an expectile state-value function and avoids querying actions from the learned policy in the critic target. In contrast, Cal-QL \citep{nakamoto2023calql} follows a more standard actor-critic formulation and learns conservative, calibrated action-value estimates for offline-to-online fine-tuning. Its TD targets depend on the current policy through next-state actions. \method follows the policy-independent direction of IQL for critic learning: TD targets are formed using a state-value estimate rather than querying the current policy. This is particularly important for \diffpols from a computational perspective, since each policy query requires several denoising steps. However, unlike IQL, we do not extract the policy through advantage-weighted BC. Instead, we use MPC to generate planner-improved trajectories and train the \diffpol with a standard supervised denoising loss. Besides, we draw inspiration from RLPD \citep{ball2023rlpd} by simply mixing offline and online samples during training, which has been shown to outperform offline-then-online training.

\section{Background}

Our work builds on RL, MPC, sampling-based trajectory optimization, and \diffpols.

\paragraph{Reinforcement learning.}
We consider a sequential decision-making problem formulated as a Markov Decision Process (MDP), defined by the tuple
$(\mathcal{S}, \mathcal{A}, \mathcal{T}, \mathcal{R}, \gamma, \rho_0)$, where $\mathcal{S}$ denotes the state space, $\mathcal{A}$ the action space, $\mathcal{T}(s'|s,a)$ the transition dynamics, $\mathcal{R}(s,a)$ the reward function, $\gamma \in [0,1]$ the discount factor, and $\rho_0$ the initial state distribution.
The objective in RL is to maximize the expected return $\sum_{t=0}^{H} \gamma^{t} r_t $. 
In model-free RL, the agent directly learns a policy
$\pi_\theta : \mathcal{S} \rightarrow \mathcal{A}$ that maximizes the expected return. In contrast, an MBRL agent learns a predictive model of the environment dynamics, $\hat{\mathcal{T}}_\theta$, and uses it for planning. Given the current state $s_t$, the agent searches for a sequence of future actions $A=\{a_t,\ldots,a_{t+H-1}\}$ over a finite horizon $H$ by rolling out candidate sequences through the learned model and selecting the one with the highest predicted return.

\paragraph{Sampling-based trajectory optimization.}
Since learned models are imperfect, prediction errors can compound over long horizons. A common solution is to solve this finite-horizon optimization problem in a receding-horizon manner: at each environment step, the agent plans over horizon $H$, executes only the first action or the first few actions, observes the new state, and replans. This procedure is known as Model Predictive Control (MPC).

For non-linear dynamics and rewards, the finite-horizon planning problem cannot be solved analytically. Sampling-based trajectory optimizers such as MPPI are therefore commonly used in MPC. At each planning step, MPPI samples a population of candidate action sequences from a proposal distribution, rolls them out through the learned dynamics model, and scores them according to their predicted return. The sampling distribution is then shifted toward high-return candidates, and this process is repeated for a small number of iterations. After optimization, the first action or the first few actions of the best sequence are executed before replanning. The planning budget is proportional to the number of sampled trajectories times the number of optimization iterations, and is therefore a key factor controlling inference time. More details about MPPI can be found in Appendix~\ref{app:mppi-details}.

\paragraph{Hybrid model-based planners.}
Hybrid planners combine model-based trajectory optimization with prior policies. Instead of sampling all candidate action sequences from a generic Gaussian distribution, they use a policy to propose informative trajectories and complement them with random samples for exploration. This biases planning toward task-relevant regions of the action space and can reduce the number of samples required for effective optimization.

To account for rewards beyond a short MPC horizon, hybrid planners often augment the predicted finite-horizon return with a terminal value estimate. A common choice is a policy-dependent terminal estimate such as $Q_\phi(s_{t+H}, \pi(s_{t+H}))$. With this terminal estimate, the planner can reason over shorter horizons while still approximating long-term returns. The optimal sequence of actions $A^*$ is thus defined as

\begin{equation}
\begin{aligned}
\label{eq:hybrid_optimization}
A^* =
\arg\max_{A \in \mathcal{A}^H}
\Bigg(
\sum_{h=0}^{H-1}
\gamma^h
R(s_{t_0+h}, a_{t_0+h})
+
\gamma^H
Q_\phi(s_{t_0+H}, \pi(s_{t_0+H}))
\Bigg), \\ 
\text{s.t.} \quad
s_{t_0+h+1}
=
\hat{\mathcal{T}}_\theta(s_{t_0+h}, a_{t_0+h}).
\end{aligned}
\end{equation}

\paragraph{Diffusion policies.}
Diffusion Policies have recently emerged as a powerful framework for imitation learning and robotic control \citep{chi2023diffusion}. Instead of representing the policy as a unimodal Gaussian distribution, \diffpols model the conditional distribution over action sequences through a denoising generative process. Given an observation $s_t$, the \diffpol generates an action chunk $\mathbf{a} = (a_t, \ldots, a_{t+T_a-1})$ by starting from Gaussian noise $\mathbf{a}^K \sim \mathcal{N}(0,I)$ and iteratively denoising it during $K$ denoising steps.

At denoising step $k \in \{1,\ldots,K\}$, a neural network
$\epsilon_\theta(\mathbf{a}^k,k,s_t)$ predicts the noise component in the current noisy action sequence. The reverse diffusion process updates the action sequence as
$\mathbf{a}^{k-1}
=
\alpha_k
\left(
\mathbf{a}^{k}
-
\gamma_k
\epsilon_\theta(\mathbf{a}^{k}, k, s_t)
\right)
+
\sigma_k \mathbf{z}$,
where $\mathbf{z}\sim\mathcal{N}(0,I)$ and $\alpha_k$, $\gamma_k$, and $\sigma_k$ are determined by the noise schedule.

The denoising model is trained to predict the injected noise using the objective
\begin{equation}
\label{eq:BC}
    \mathcal{L}_{\mathrm{DP}}(\theta)
    =
    \mathbb{E}_{\mathbf{a}, k, \epsilon}
    \left[
    \left\|
    \epsilon -
    \epsilon_\theta(\mathbf{a}^k, k, s_t)
    \right\|^2
    \right].
\end{equation}

Because \diffpols can represent complex and multi-modal action distributions, they are particularly suitable for robotic tasks where several distinct strategies may solve the same problem. In the context of hybrid planning, \diffpols can serve as expressive proposal distributions, guiding sampling-based planners toward task-relevant and potentially multi-modal action regions while allowing subsequent refinement through MPC.

\section{Method: \method} 

\begin{figure}
    \centering
    \includegraphics[width=\linewidth]{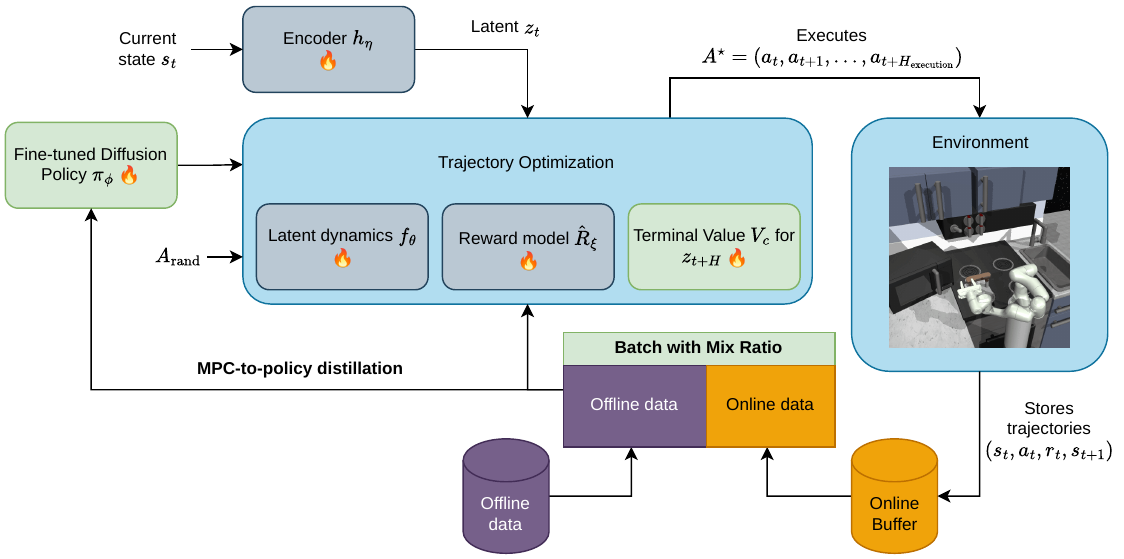}
    \caption{\textbf{Overview of the \method training process.} The agent uses a latent world model to perform short-horizon MPC planning. The fine-tuned \diffpol acts as a proposal prior. The resulting planner-improved trajectories are stored in an online buffer and mixed with expert offline data. This data mixture is then used to distill planned trajectories back into the \diffpol, creating an iterative improvement loop.}
    \label{fig:MODIP_training}
\end{figure}

Building on the hybrid planning formulation introduced above, \method uses MPC as a policy improvement operator for \diffpols. Instead of directly optimizing the \diffpol with an RL objective, a model-based planner first improves candidate action sequences, and the resulting trajectories are then distilled back into the \diffpol using a standard denoising BC loss.

As illustrated in \cref{fig:MODIP_training}, \method consists of five learned components. As in the TOLD model of TD-MPC \citep{Hansen2022tdmpc}, an encoder $h_\eta$ maps observations to latent states $z_t=h_\eta(s_t)$, a latent dynamics model $f_\theta$ predicts future latent states under candidate actions, and a reward model $\hat R_\xi$ scores short-horizon imagined rollouts. Departing from TD-MPC, a terminal state-value function $V_\psi$ estimates long-horizon returns beyond the MPC horizon and a \diffpol $\pi_\phi$ serves as an expressive action sequence prior for the planner.
The model components used in MPC are trained with the joint objective:
\begin{equation}
\label{eq:modip-total-model-loss}
\mathcal{L}_{\mathrm{model}} =
\lambda_{\mathrm{cons}}\mathcal{L}_{\mathrm{cons}}
+
\lambda_{\mathrm{rew}}\mathcal{L}_{\mathrm{rew}}
+
\lambda_{\mathrm{value}}
\left(
\mathcal{L}_{Q}
+
\mathcal{L}_{V,A}
\right),
\end{equation}
where the individual losses are defined in the following subsections.

Training proceeds in two stages. First, the \diffpol and all MPC components are trained on the offline dataset. Second, during online fine-tuning, actions are selected by the DP-guided MPC planner, and the resulting trajectories are stored in an online replay buffer. We follow the mixing strategy of RLPD \citep{ball2023rlpd}, updates are performed on minibatches that combine offline and online data. In contrast to a fixed mixing ratio, \method uses a time-varying ratio $\beta(t)$  (see Appendix \ref{app:hyperparams}) that gradually increases the fraction of online samples, keeping early updates anchored to the offline data while gradually shifting toward improved trajectories.

Three design choices make the \method loop efficient for \diffpols. First, MPC-to-policy distillation improves the \diffpol without changing its supervised denoising objective: the policy is trained by BC on trajectories generated by a stronger MPC planner. Second, \method uses a terminal state value $V(z)$ instead of the common estimate $Q(z,\pi(z))$, avoiding costly \diffpol queries at terminal states during planning. Third, critic learning is made policy-independent by forming TD targets from state-value estimates rather than actions sampled from the current policy. This avoids repeated policy denoising iterations during critic updates and reduces training cost.

The remainder of this section details the three main parts of \method: MPC-to-policy distillation for \diffpol improvement, DP-guided MPC with a terminal state value, and the training of the latent world model and policy-independent critic.

\subsection{MPC-to-policy distillation as diffusion policy improvement}

The core training mechanism of \method is MPC-to-policy distillation. After pretraining the \diffpol on the offline dataset with BC, during online fine-tuning, actions are selected by the DP-guided MPC planner, and the resulting trajectories are stored in an online replay buffer.  The \diffpol is then updated with the same denoising BC objective:
\begin{equation}
\mathcal{L}_{\mathrm{DP}}(\phi)
=
\mathbb{E}_{\mathbf a,k,\epsilon}
\left[
\left\|
\epsilon -
\epsilon_\phi(\mathbf a^k,k,h_\eta(s_t))
\right\|_2^2
\right].
\end{equation}
Unlike ordinary BC, the target action sequences $\mathbf a$ in \cref{eq:BC} come from trajectories generated  by the MPC planner. This procedure distills the planner behavior into the \diffpol that improves through the data distribution rather than through a change in the diffusion objective.

When a pretrained policy is available, we use its action $\mathbf a_{reg}$ to regularize policy learning:
\begin{equation}
\label{eq:modip-teacher-loss}
\mathcal{L}_{\pi}
=
\mathcal{L}_{\mathrm{DP}}(\mathbf a_{\mathrm{MPC}})
+
\lambda(t)
\mathcal{L}_{\mathrm{DP}}(\mathbf a_{reg}),
\end{equation}
where $\lambda(t)$ is annealed during online training. This stabilizes the transition from offline imitation to planner-guided improvement and prevents the \diffpol from drifting too abruptly away from the initial behavioral prior.

\subsection{Inference-efficient hybrid planning with terminal state value}
\label{subsec:modip-planner}
To generate improved trajectories, the DP-guided MPC planner optimizes action sequences in the learned latent space. A key design choice is to use a terminal state-value function $V(z)$ instead of the standard terminal state-action value $Q(z,\pi(z))$.
~Indeed, we need to evaluate the terminal estimate for many candidate trajectories at every planning step: when $\pi$ is a \diffpol, using $Q(z,\pi(z))$ would require running the multi-step denoising process at each terminal state to obtain $\pi(\hat z_{t+\Hmpc})$. In contrast, $V_\psi(\hat z_{t+\Hmpc})$ scores the final latent state with a single forward pass, reducing inference cost while accounting for returns beyond the short MPC horizon. Given the current latent state $z_t$, the DP-guided MPC planner approximately solves the following optimization problem with MPPI:

\begin{equation}
\label{eq:plan}
\begin{aligned}
A^*
=
\arg\max_{A \in \mathcal{A}^{\Hmpc}}
\left[
\sum_{k=0}^{\Hmpc-1}
\gamma^k
\hat R_\xi(\hat z_{t+k}, a_{t+k})
+
\gamma^{\Hmpc}
V_\psi(\hat z_{t+\Hmpc})
\right],
\begin{cases}
\hat z_t &= h_\eta(s_t),\\
\hat z_{t+k+1} &= f_\theta(\hat z_{t+k},a_{t+k}).
\end{cases}
\end{aligned}
\end{equation}
MPPI samples candidate action sequences $A=\{a_t,\ldots,a_{t+\Hmpc-1}\}$ from a hybrid proposal: some are generated by the \diffpol, providing high-likelihood action sequences, while the rest are sampled from a Gaussian proposal. Candidate sequences are rolled out in the latent model, scored with \cref{eq:plan}. This combines structured proposals from the \diffpol with local exploration and model-based refinement.

\subsection{Latent world model}

The DP-guided MPC planner defined in \cref{subsec:modip-planner} uses a latent world model composed of an encoder $h_\eta$, a latent dynamics $f_\theta$, and a reward model $\hat R_\xi$. Given a sequence from the replay buffer, \method unrolls the dynamics for $\Hmpc$ steps and trains the predictions to match target latent states encoded by an exponential moving average (EMA) target encoder $h_{\bar\eta}$. In offline RL, this model must remain accurate over the full planning horizon to reduce the risk of model exploitation and value overestimation during planning. The consistency loss is:

\begin{equation}
\label{eq:modip-consistency-loss}
\mathcal{L}_{\mathrm{cons}}(\eta,\theta) =
\sum_{h=0}^{\Hmpc-1}
\lambda_{\mathrm{roll}}^h
\left\|
f_\theta(\hat z_{t+h}, a_{t+h})
- h_{\bar\eta}(s_{t+h+1})
\right\|_2^2,
\end{equation}
where $\lambda_{\mathrm{roll}}\in(0,1]$ weights errors along the rollout and is chosen close to $1$ so that later prediction errors are not overly down-weighted.

In parallel, a parametric reward model $\hat R_\xi(\hat z_t,a_t)$ learns to predict the immediate task reward. 
For dense-reward tasks, we train it by regression to the scalar reward. For sparse-reward manipulation tasks, we train the reward model as a success classifier. Let $y^r_t=\mathbbm{1}[r_t>0]$ and let $\ell_\xi$ denote the reward logit. The reward loss is:
\begin{equation}
\label{eq:modip-reward-loss}
\mathcal{L}_{\mathrm{rew}}(\xi)
=
\sum_{h=0}^{\Hmpc-1}
\lambda_{\mathrm{roll}}^h
\begin{cases}
\mathrm{MSE}\left(
\hat R_\xi(\hat z_{t+h},a_{t+h}) , r_{t+h}
\right)
& \text{for dense rewards},\\[4pt]
\mathrm{BCE}\left(
\ell_\xi(\hat z_{t+h},a_{t+h}),
y^r_{t+h}
\right)
& \text{for sparse rewards}.
\end{cases}
\end{equation}

\subsection{Policy-independent critic learning}
\label{subsec:policy-independent-critic}

In standard actor--critic methods, TD targets often depend on the current policy through terms such as $Q(s',\pi_\theta(s'))$ \citep{nakamoto2023calql,mark2024policy}. This coupling is undesirable for \diffpols: querying the policy is expensive because action generation requires several denoising steps, and policy errors can propagate into the critic. We therefore follow the spirit of AFU \citep{perrin2024afu} and form \emph{policy-independent} targets directly from replay. For a transition $(s_t,a_t,r_t,s_{t+1},d_t)$ and $z_{t+1} = h_{\bar\eta}(s_{t+1})$, we replace the usual policy-dependent target  with a target state-value estimate $V_\psi(z_{t+1}))$ computed using slowly updated EMA networks:

\begin{equation}
\label{eq:modip-v-target}
y_t
= r_t + \gamma (1-d_t)
\min_{j\in\{1,2\}} V_{\bar\psi_j}\left(h_{\bar\eta}(s_{t+1})\right),
\end{equation}
where $\bar\eta$ and $\bar\psi_j$ denote EMA target parameters of the encoder and value networks, respectively. 

The action-value function is then trained by regression:
\begin{equation}
\label{eq:modip-q-loss}
\mathcal{L}_{Q}(\omega) =
\mathbb{E}_{(s,a,r,s')\sim \mathcal{B}}
\left[
\left(Q_\omega(h_\eta(s),a)-y_t\right)^2
\right].
\end{equation}

In addition to $Q_\omega$, we train two value heads $V_{\psi_i}$ and two advantage heads $A_{\zeta_i}$, for $i\in\{1,2\}$. Following AFU \citep{perrin2024afu}, the value--advantage decomposition is trained to recover state-value estimates without querying the current policy. The resulting value heads provide the terminal value used by the planner in \cref{eq:plan}, while the advantage heads capture action-dependent deviations from the state value. We provide the full value--advantage objective in Appendix~\ref{app:va-objective}.

\section{Results}
Our experiments evaluate four aspects of \method. 
First, we compare \method to methods that directly apply RL to \diffpols, to assess whether MPC-to-policy distillation provides a competitive alternative to direct RL fine-tuning. Second, we test whether \method improves a pretrained \diffpol beyond its offline BC initialization. Third, we study whether using a \diffpol as an MPC proposal prior is more effective than relying on a simple MLP policy. Finally, we analyze the impact of terminal state-value formulation and policy-independent critic learning on computational efficiency and performance.


\paragraph{Benchmarks.}
Evaluations are performed on three benchmark families: D4RL MuJoCo locomotion, D4RL Kitchen, and Robomimic state-based manipulation. D4RL MuJoCo includes standard dense-reward continuous-control tasks such as HalfCheetah, Hopper, and Walker2d. D4RL Kitchen evaluates long-horizon manipulation, where the agent must compose multiple subtasks over extended horizons. Robomimic evaluates precise multi-stage manipulation from low-dimensional state observations, including tasks such as Lift, Can and Square. We report the cumulative return for MuJoCo, and task success rates for Kitchen and Robomimic. To ensure a fair comparison, all methods are evaluated under a fixed computational budget.

\paragraph{Baselines.}
We compare \method against a set of baselines chosen to reflect the main directions in recent work on improving \diffpols for control. BC denotes the pretrained behavior-cloned diffusion policy. DQL \citep{wang2022diffusion} combines diffusion policies with offline Q-learning by augmenting the denoising objective with value maximization.
DPPO \citep{ren2025diffusion} applies policy-gradient to fine-tune \diffpols with RL. PA-RL \citep{mark2024policy} improves expressive policy classes, including \diffpols, through critic-guided action optimization followed by supervised policy improvement. DSRL \citep{wagenmaker2025steering} adapts a pretrained \diffpol by performing RL in its latent noise space rather than directly optimizing the policy parameters. Finally, TD-MPC2 \citep{hansen2024td} is a strong hybrid planning baseline: it combines a learned world model with trajectory optimization, but relies on a Gaussian policy instead of a \diffpol. For a fair comparison, we train TD-MPC2 with the same offline pretraining procedure used for \method. See Appendix \ref{app:implementation-details} for implementation details.

\subsection{Main Results}
\label{subsec:main-results}

\begin{table}[ht]
\caption{Final performance of \method against \diffpol fine-tuning methods and model-based baselines. 
For MuJoCo we report returns, for Kitchen and for RoboMimic we report success rates. Results in italics are reported from \citep{wagenmaker2025steering}.
}
\centering
\small
\setlength{\tabcolsep}{5pt}
\renewcommand{\arraystretch}{1.15}
\resizebox{\textwidth}{!}{
\begin{tabular}{llccccccc}
\toprule
\textbf{Benchmark} & \textbf{Task} & \textbf{BC} &\textbf{DQL} & \textbf{DSRL} & \textbf{DPPO} & \textbf{PA-RL} & \textbf{TD-MPC2} & \textbf{\method (ours)} \\
\midrule

\multirow{3}{*}{D4RL/MuJoCo}
&  halfcheetah & 5108 $\pm$ 730 &\textit{88 $\pm$ 366} & \textit{4670 $\pm$ 59} & 4290 $\pm$ 35 & \textbf{14254 $\pm$ 1564} & 12054 $\pm$ 1336 & 13775 $\pm$ 203 \\
& walker & 5721 $\pm$ 324 & \textit{271 $\pm$ 94}
& \textit{4064 $\pm$ 66 } & 2848 $\pm$ 86 & 5361 $\pm$ 466 & 5433 $\pm$ 93 & \textbf{6081 $\pm$ 66} \\
& hopper & \textbf{3050 $\pm$ 90} & \textit{360 $\pm$ 53}
& \textit{3023 $\pm$ 32} & 1612 $\pm$ 23 & 2855 $\pm$ 436 & 2129 $\pm$ 316 & \textbf{3281 $\pm$ 370}\\
\midrule

\multirow{2}{*}{D4RL/Kitchen}

& Complete & 0.41 $\pm$ 0.20 
 & \textit{0.42 $\pm$ 0.05}
 & \xmark & 0.88 $\pm$ 0.02 & 0.85 $\pm$ 0.01
  & 0.65 $\pm$ 0.13 & \textbf{0.94 $\pm$ 0.14}\\
 & Partial & 0.32 $\pm$ 0.06
 & \textit{0.45 $\pm$ 0.03} & \xmark & 0.67 $\pm$ 0.04 & 0.93 $\pm$ 0.01  & 0.55 $\pm$ 0.28 & \textbf{0.98 $\pm$ 0.01}\\
\midrule

\multirow{3}{*}{RoboMimic }
& Lift & 0.95 $\pm$ 0.01 & \textit{0.51 $\pm$ 0.23}& \textit{\textbf{1.0 $\pm$ 0.0}} & \textbf{0.99 $\pm$ 0.01} & \textbf{0.99 $\pm$ 0.02} & 0.0 & \textbf{0.98} $\pm$ 0.03 \\
& Can & 0.87 $\pm$ 0.04
 & \textit{0.0} & \textit{0.94 $\pm$ 0.02} & 0.93 $\pm$ 0.02 & 0.90 $\pm$ 0.02 & 0.0 & 0.92 $\pm$ 0.01 \\
& Square & 0.62 $\pm$ 0.09
 & \textit{0.0}  &\textit{\textbf{ 0.90 $\pm$ 0.01}}& 0.73 $\pm$ 0.01 &  0.81 $\pm$ 0.03 & 0.0 & 0.85 $\pm$ 0.08\\
\midrule

\end{tabular}
}
\label{tab:main_results}
\end{table}

Table~\ref{tab:main_results} reports the final performance of \method against \diffpol fine-tuning methods and hybrid planning baselines. Overall, \method achieves strong performance across dense-reward locomotion, long-horizon manipulation, and state-based robotic manipulation.

On D4RL MuJoCo, \method consistently outperforms TD-MPC2 across all three tasks, suggesting that using a \diffpol as an expressive action sequence prior can improve over an MLP policy prior in hybrid planning. Notably, \method and PA-RL largely outperform the other \diffpol baselines on HalfCheetah. We hypothesize that this gap is partly due to the need for frequent feedback in this task: methods that execute long action chunks generated by a \diffpol may struggle to quickly recover from prediction or control errors. In contrast, PA-RL optimizes single-step actions with a critic, while \method replans with MPC at each control step. Therefore, both approaches enable more reactive control, which may explain their stronger performance on this task.\\
On D4RL Kitchen, \method achieves the best performance on both Complete and Partial tasks. These tasks require long-horizon coordination and sparse reward discovery, making them well-suited to the hybrid planning mechanism of \method. The results indicate that DP-guided MPC can generate stronger trajectories than directly fine-tuning \diffpols with RL, and that these trajectories provide useful supervision for policy improvement. The learning curves in \cref{fig:mosaic} in Appendix~\ref{app:curves} further show the online improvement behavior of \method.\\
On RoboMimic, \method is competitive with the strongest \diffpol baselines. It achieves near-saturated performance on Lift, strong performance on Can, and improves over DPPO and PA-RL on Square, while DSRL obtains the highest score.\\
These results suggest that \method is particularly effective in settings where model-based refinement and \diffpol priors are complementary, while direct \diffpol adaptation methods can remain strong when the pretrained diffusion prior is already highly aligned with the task.

\begin{table}[ht]
\caption{
Ablations of the main components of \method. The first row reports the full \method method. 
Each subsequent row changes one component while keeping the others fixed.
}
\centering
\small
\setlength{\tabcolsep}{4pt}
\renewcommand{\arraystretch}{1.12}
\resizebox{\textwidth}{!}{
\begin{tabular}{llrrrrr}
\toprule
\textbf{Ablation} & \textbf{Variant} 
& \textbf{HalfCheetah} 
& \textbf{Walker2d} 
& \textbf{Hopper} 
& \textbf{Kitchen-Complete} 
& \textbf{Kitchen-Partial} \\
\midrule

\multirow{1}{*}{Full method}
& \method 
& \textbf{13775 $\pm$ 203} 
& \textbf{6081 $\pm$ 66} 
& \textbf{3115 $\pm$ 23} 
& \textbf{0.94 $\pm$ 0.14} 
& \textbf{0.98 $\pm$ 0.01} \\
\midrule

\multirow{1}{*}{Hybrid planning}
& Policy-only 
& 1178.7 $\pm$ 157 
& \textbf{6044 $\pm$ 24 }
& \textbf{3074 $\pm$ 12 }
& 0.77 $\pm$ 0.03
& 0.74 $\pm$ 0.05 \\

\multirow{1}{*}{Terminal estimate}
& \method with $Q(s,\pi(s))$ 
& 7113 $\pm$ 514 
& 3278 $\pm$ 98 
& 1483 $\pm$ 74 
& 0.55 $\pm$ 0.10 
& 0.60 $\pm$ 0.18 \\

\multirow{1}{*}{Policy prior}
& MLP policy 
& 12413 $\pm$ 862  
& 4948 $\pm$ 173 
& 1815 $\pm$ 242 
& 0.45 $\pm$ 0.74 
& 0.52 $\pm$ 0.83  \\

\multirow{1}{*}{Offline initialization}
& Without offline pretraining 
& 11736 $\pm$ 319
& 5102 $\pm$ 149
& 1255 $\pm$ 95
& 0.74 $\pm$ 0.06
& \textbf{0.98 $\pm$ 0.01} \\
\bottomrule
\end{tabular}
}
\label{tab:ablation_components}
\end{table}

\subsection{Analysis and Ablations}
\label{subsec:analysis-ablations}

We ablate the main design choices of \method to understand which components are responsible for the observed gains. Table~\ref{tab:ablation_components} summarizes the component ablations. The first row reports the full method, while each subsequent row changes one component and keeps the others fixed.

\paragraph{MPC-to-policy distillation and hybrid planning.}
Table~\ref{tab:ablation_components} compares the full \method planner against a policy-only variant that executes the \diffpol without MPC refinement. The full planner substantially improves over the policy-only controller on HalfCheetah and both Kitchen tasks, showing that model-based refinement is especially beneficial when the policy alone is insufficient. On Walker2d and Hopper, the policy-only variant is already strong and performs close to the full planner. This supports the MPC-to-policy distillation view of \method: online trajectories used to fine-tune the \diffpol are generated by a stronger planner which improves the \diffpol performance, while model-based planning can further improve task performance in several challenging settings.

\paragraph{Choice of terminal value for inference efficiency.}
We evaluate the effect of using $V(z)$ instead of the policy-dependent estimate $Q(z,\pi(z))$ as terminal value. Table~\ref{tab:ablation_components} shows that $V(z)$ improves performance across all evaluated tasks, indicating that it provides a more effective value signal for \diffpol-guided planning. In addition, Table~\ref{tab:inference_time_terminal_value} shows that $V(z)$ reduces inference time from $85.75$ to $29.54$ seconds per $10^3$ environment steps, corresponding to a $2.90\times$ speedup. These results show that the state-value formulation improves both planning efficiency and final performance.

\begin{table}[ht]
\caption{
Inference-time comparison on HalfCheetah between terminal $Q(z,\pi(z))$ and terminal $V(z)$. 
Unlike $V(z)$, $Q(z,\pi(z))$ requires querying the multi-step \diffpol denoising process at terminal states, increasing inference time.
}
\centering
\small
\setlength{\tabcolsep}{4pt}
\begin{tabular}{lcc}
\toprule
\textbf{Terminal estimate} 
&  $Q(z,\pi(z))$ & $V(z)$ (ours) 
 \\
\midrule
\textbf{Inference time (seconds / $10^3$ environment steps)}  
& 85.75 & 29.54
\\
\bottomrule
\end{tabular}
\label{tab:inference_time_terminal_value}
\end{table}
\paragraph{Policy-independent critic learning for better training time.}
We evaluate the effect of decoupling critic learning from the current \diffpol. A policy-coupled TD target requires querying the \diffpol to compute next-state actions, which is expensive because each action is produced through a multi-step denoising process.  In contrast, our policy-independent value target avoids this cost. 
As shown in \cref{tab:ablation_critic_learning}, the policy-independent formulation substantially reduces the cost of MPC pretraining and online fine-tuning. Overall, policy-independent targets reduce total training time from $\sim40$h to $\sim25$h, corresponding to a $\sim1.6\times$ speedup. This confirms that decoupling critic learning from the \diffpol is crucial to make DP-guided MPC computationally efficient.

\begin{table}[ht]
\caption{Ablation on critic learning efficiency on HalfCheetah. We compare policy-independent targets against standard policy-coupled TD targets. Times are reported in hours and minutes.}
\centering
\small
\setlength{\tabcolsep}{4pt}
\begin{tabular}{lccccc}
\toprule
\textbf{Variant} & \textbf{\diffpol pretraining} & \textbf{MPC pretraining} & \textbf{Data collection}  & \textbf{Fine-tuning} & \textbf{Total} \\
\midrule
Policy-independent targets 
& 2h 13m 
& \textbf{4h 52m} 
& 3h 11m 
& \textbf{15h 00m} 
& \textbf{25h 16m} \\
Policy-coupled TD targets  
& 2h 13m 
& 10h 09m 
& 3h 11m 
& 24h 43m 
& 40h 16m \\
\bottomrule
\end{tabular}
\label{tab:ablation_critic_learning}
\end{table}

\paragraph{Offline initialization.}
We evaluate the effect of offline pretraining the latent dynamics model, reward model, and critic before online fine-tuning. Since the planner generates the trajectories used to improve the \diffpol, its quality early in training is important. 
Table~\ref{tab:ablation_components} shows that offline initialization improves performance on MuJoCo and Kitchen Complete, with particularly large gains on Hopper. On Kitchen Partial, both variants reach similar final performance, suggesting that offline initialization is most beneficial when early planner quality strongly affects exploration and trajectory collection. Overall, these results indicate that offline pretraining provides a useful planner from the beginning of online interaction, improving data collection and subsequent policy distillation.

\paragraph{Policy prior parameterization.}
We compare DP-guided MPC against a planner guided by a Gaussian MLP policy. As shown in Table~\ref{tab:ablation_components}, the \diffpol prior improves performance on all tasks, with especially large gains on Hopper and Kitchen. This supports the hypothesis that the benefit of \method comes not only from using a learned prior in MPC, but from using an expressive \diffpol prior that captures richer action sequence distributions.

\section{Conclusion}

In this work we presented \method, a hybrid planning algorithm which uses a diffusion policy as MPC prior and learns a policy-independent critic to evaluate trajectories beyond the MPC horizon. \method benefits from the rich distribution modelling capabilities of \diffpols during the offline pretraining stage, and is competitive with both \diffpol fine-tuning methods based on RL and hybrid planning methods relying on standard MLP policies. We also showed that, when using a \diffpol as policy prior, using a value function $V$ instead of an action-value function matters a lot, as it significantly reduces inference time and improves performance.
A limitation of our work is that, although diffusion policies have shown a high versatility in learning from pixels, all the benchmarks studied here use real-valued vectors as state information. Assessing the potential of \method to learn from images is an obvious next step for future work.

\bibliography{references}
\bibliographystyle{ewrl_2026}

\newpage
\appendix
\section{MPPI Planner Details}
\label{app:mppi-details}

At each planning step, MPPI maintains a proposal distribution over action sequences, typically a Gaussian distribution $A\sim\mathcal{N}(\mu,\sigma^2)$. It samples $N$ candidate action sequences
$A^n=\{a^{(n)}_t,\ldots,a^{(n)}_{t+H-1}\}$ and evaluates each sequence by rolling it out with the learned dynamics model. Let $R_n$ denote the predicted return of candidate sequence $A^n$.

After scoring the candidate trajectories, the proposal distribution is updated toward high-return sequences. In our implementation, we select the top $K$ trajectories, called elites, and weight them using an exponential transformation of their returns:
\[
w_k =
\frac{\exp(\tau R_k)}
{\sum_{j=1}^{K}\exp(\tau R_j)},
\]
where $\tau$ is an inverse-temperature parameter. The proposal mean and variance are then updated as
\[
\mu \leftarrow \sum_{k=1}^{K} w_k A^k,
\qquad
\sigma^2 \leftarrow
\sum_{k=1}^{K} w_k (A^k-\mu)^2.
\]
This update is repeated for $I$ optimization iterations. The final mean sequence $\mu_I$ is used as the optimized plan, and the first action, or first few actions, are executed before replanning at the next environment step. We define the planning budget as $N\times I$, which is the dominant factor controlling inference time in sampling-based MPC.

\section{Value--Advantage Objective}
\label{app:va-objective}

\begin{figure}[htbp]
    \centering
    \includegraphics[width=\linewidth]{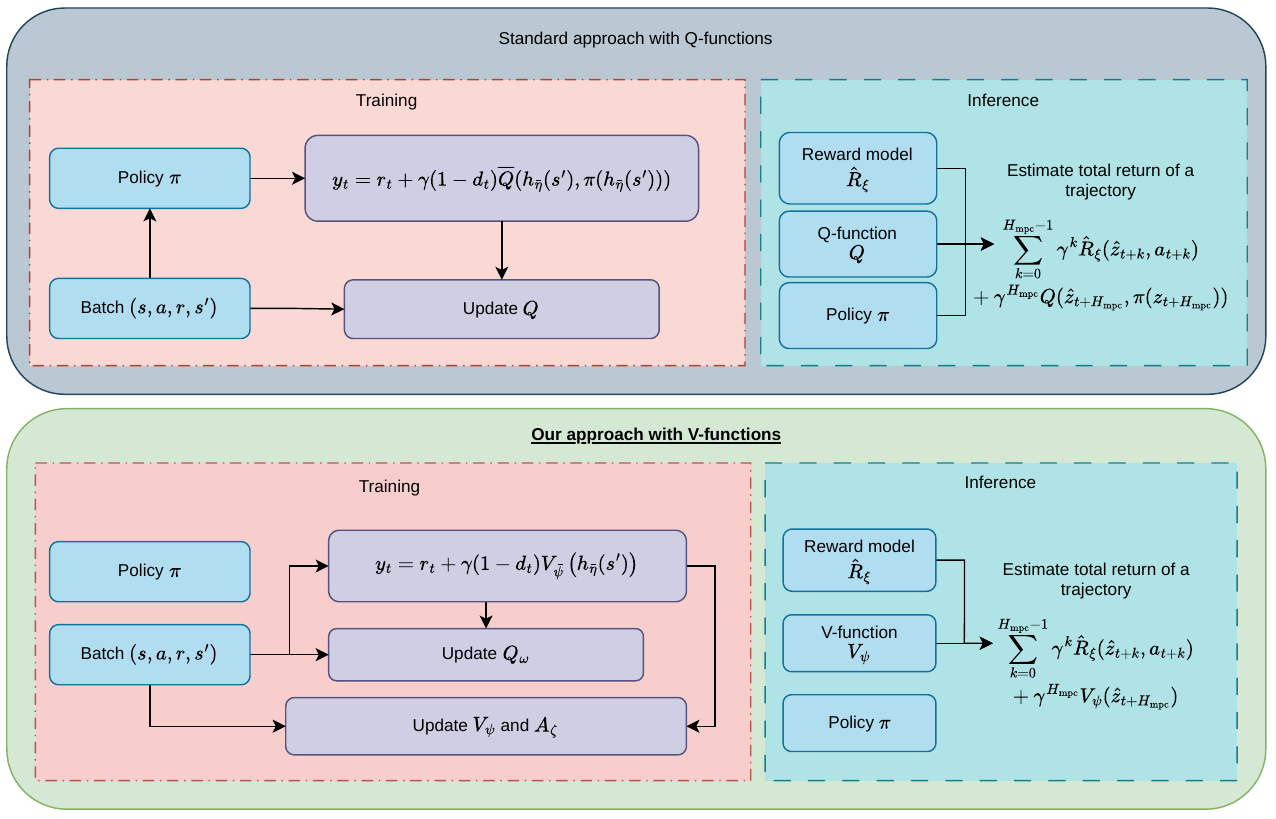}
    \caption{\textbf{Comparison between the standard Q-function approach (top) and our proposed policy-independent V-function approach (bottom).} In the standard approach, the TD target $y_t$ depends on the current policy $\pi$, necessitating expensive \diffpol queries and coupling critic errors to policy performance. In our approach, we decouple the critic from the policy by learning state-value ($V$) and advantage ($A$) functions directly from the replay buffer. The TD target $y_t$ is computed using target V-functions, eliminating the need to query the policy during training and inference}
    \label{fig:V-function}
\end{figure}

This appendix provides the full value--advantage objective used to train the terminal state-value function. In addition to the action-value function $Q_\omega$, we train two value heads $V_{\psi_i}$ and two advantage heads $A_{\zeta_i}$, for $i\in\{1,2\}$. The goal is to learn a policy-independent state-value estimate without querying the current policy. Let $z_t=h_\eta(s_t)$ and let $y_t$ denote the policy-independent TD target defined in \cref{eq:modip-v-target}. For each head $i$, following \cite{perrin2024afu} we define
\begin{equation}
\label{eq:modip-va-loss}
\mathcal{L}_{V,A}^{(i)}(\psi_i,\zeta_i)
=
\mathbb{E}_{(s_t,a_t)\sim\mathcal{B}}
\left[
Z\left(
\Upsilon_i(z_t,a_t)-y_t,
A_{\zeta_i}(z_t,a_t)
\right)
\right],
\end{equation}
where
\begin{equation}
\label{eq:modip-z-loss}
Z(x,y)
=
\begin{cases}
(x+y)^2, & \text{if } x \geq 0,\\
x^2+y^2, & \text{otherwise.}
\end{cases}
\end{equation}

The term $\Upsilon_i$ controls how gradients are propagated through the value estimate:
\begin{equation}
\label{eq:modip-upsilon}
\Upsilon_i(z_t,a_t)
=
(1-\varrho I_i^{z_t,a_t})V_{\psi_i}(z_t)
+
\varrho I_i^{z_t,a_t}
\operatorname{sg}\!\left[V_{\psi_i}(z_t)\right],
\end{equation}
where $\operatorname{sg}[\cdot]$ denotes the stop-gradient operator and $\varrho\in[0,1]$ controls the amount of gradient reduction applied to the value estimate. The indicator $I_i^{z_t,a_t}$ determines when this gradient reduction is applied:
\begin{equation}
\label{eq:modip-indicator}
I_i^{z_t,a_t}
=
\mathbbm{1}
\left[
V_{\psi_i}(z_t)+A_{\zeta_i}(z_t,a_t)
<
Q_\omega(z_t,a_t)
\right].
\end{equation}

The final value--advantage loss averages over the two heads:
\begin{equation}
\label{eq:modip-va-total}
\mathcal{L}_{V,A}
=
\frac{1}{2}
\sum_{i=1}^{2}
\mathcal{L}_{V,A}^{(i)}.
\end{equation}
The learned value heads provide the terminal value used by the MPC planner, while the advantage heads model action-dependent deviations from the state value.

\section{Implementation Details}
\label{app:implementation-details} 

\paragraph{Diffusion policy architecture.}
We implement the \diffpol as an MLP-based conditional diffusion model over action chunks such as in IDQL \citep{hansen2023idql}. Given a latent state $z_t=h_\eta(s_t)$, the policy generates an action sequence $\mathbf a=(a_t,\ldots,a_{t+H_\pi-1})$ through a DDPM denoising process. Following the standard $\epsilon$-prediction parameterization, the denoising network predicts the noise component $\epsilon_\phi(\mathbf a^k,k,z_t)$ at denoising step $k$. The input to the score network is the concatenation of the flattened noisy action sequence, the latent state condition, and a sinusoidal embedding of the denoising timestep. The score network is a LN\_Resnet \citep{hansen2023idql} composed of three residual MLP blocks. The output is reshaped to the action chunk dimension and represents the predicted diffusion noise. We use a squared-cosine DDPM noise schedule with $K$ denoising steps.

\paragraph{Latent world model.}
The MPC planner uses a TOLD-style latent world model composed of an encoder $h_\eta$, latent dynamics $f_\theta$, and reward model $\hat R_\xi$. For state-based observations, the encoder is an MLP that maps observations to a latent state. The dynamics model takes the concatenated latent-action pair $(z_t,a_t)$ and predicts the next latent state $z_{t+1}$. Both the encoder and dynamics use MLP blocks with LayerNorm and Mish activations, and the dynamics output uses SimNorm normalization. The reward model also takes $(z_t,a_t)$ as input and outputs either a scalar reward prediction for dense-reward tasks or a reward logit for sparse-reward tasks. In sparse settings, the logit is converted into a success probability with a sigmoid and used by the planner as the predicted reward.

\paragraph{Critic and value networks.}
The critic module contains an action-value network $Q_\omega(z,a)$, two state-value networks $V_{\psi_1}(z),V_{\psi_2}(z)$, and two advantage networks $A_{\zeta_1}(z,a),A_{\zeta_2}(z,a)$. The $Q$ and advantage networks take the latent-action pair as input, while the value networks only take the latent state. All heads are implemented as MLPs with LayerNorm and Mish activations. The final layers of the $Q$, value, and advantage heads are initialized to zero, following TD-MPC-style initialization. The value heads provide the terminal value used by the MPC planner, while the advantage heads are used only during critic training.

\paragraph{Target networks.}
We maintain exponential moving average (EMA) target networks for the encoder and value functions. The target encoder is used to compute latent targets for the dynamics consistency loss, while the target value networks are used to compute policy-independent TD targets. After each gradient update, target parameters are updated from online parameters using EMA.

\paragraph{Optimizer and training.}
All networks are trained with Adam. Training proceeds in two stages. First, the diffusion policy, latent world model, reward model, and critic are pretrained on the offline dataset. Then, during online fine-tuning, trajectories collected by the DP-guided MPC planner are stored in an online replay buffer. Gradient updates are performed on minibatches which contain offline and online samples with a time-varying mixing ratio. 

\paragraph{Offline-to-online replay mixing.}
During online fine-tuning, we follow the replay-mixing principle of RLPD \citep{ball2023rlpd}, which combines offline data with newly collected online data in each training minibatch. However, instead of using a fixed mixing ratio, we use a time-varying online fraction $\beta(t)$. At online iteration $t$, a minibatch $\mathcal{B}$ is composed of samples from the online replay buffer and the offline dataset according to
\[
|\mathcal{B}_{\mathrm{on}}(t)|=\beta(t)|\mathcal{B}|,
\qquad
|\mathcal{B}_{\mathrm{off}}(t)|=(1-\beta(t))|\mathcal{B}|.
\]
We linearly increase $\beta(t)$ from an initial value $\beta_0$ to a final value $\beta_{\mathrm{final}}$ over a warm-up duration $T_\beta$:
\[
\beta(t)
=
\beta_0
+
\left(\beta_{\mathrm{final}}-\beta_0\right)
\min\left(\frac{t}{T_\beta},1\right).
\]
This keeps early updates anchored to the offline demonstrations while gradually shifting training toward planner-generated trajectories as online data become more informative.

\paragraph{Frozen BC regularization.}
During online fine-tuning, we freeze the pretrained BC diffusion policy and use it to regularize the updated policy. This constrains early policy updates to remain close to the initial behavior prior while the online buffer is still small. The policy loss is
\[
\mathcal{L}_{\pi}
=
\mathcal{L}_{\mathrm{DP}}(\mathbf a_{\mathrm{data}})
+
\lambda_{\mathrm{reg}}(t)
\mathcal{L}_{\mathrm{DP}}(\mathbf a_{\mathrm{reg}}),
\]
where $\mathbf a_{\mathrm{data}}$ is sampled from the mixed replay distribution and $\mathbf a_{\mathrm{reg}}$ is sampled from the frozen BC policy. The regularization weight is linearly annealed from $\lambda_0$ to $\lambda_{\mathrm{final}}$ over $T_\lambda$ online updates:
\[
\lambda_{\mathrm{reg}}(t)
=
\lambda_0
+
(\lambda_{\mathrm{final}}-\lambda_0)
\min\left(\frac{t}{T_\lambda},1\right).
\]
We typically choose $\lambda_{\mathrm{final}}=0$, gradually relaxing the BC constraint as planner-generated trajectories become more reliable.
\newpage
\section{Hyper-parameters}
\label{app:hyperparams}
\begin{table}[htbp]
\centering
\caption{\textbf{Hyper-parameters.} We here list hyper-parameters grouped by component category.}
\label{tab:hyperparams}
\begin{tabular}{l l l}
\toprule
\textbf{Category} & \textbf{Hyper-parameter} & \textbf{Value} \\
\midrule

\multirow{9}{*}{General}
&Learning Rate & $3 \times 10^{-4}$ \\
&Discount factor $\gamma$ & 0.99 \\
&Batch Size & 256 \\
&RLPD mix-ratio $ (\beta_0, \beta_{final}, T_\beta)$ & (0.05, 0.50, $200 000$) \\
&BC regularization $ \lambda_{reg}  (\lambda_0, \lambda_{final}, T_\lambda)$ & (1.0, 0.0, $200 000$) \\
&Optimizer & Adam \\
&Reward coefficient & 0.5 \\
&Value coefficient & 0.1 \\
&Consistency loss coefficient & 4.0 \\
&Rollout coefficient $\lambda_\text{roll}$ & 0.99 \\
&Gradient norm clipping & 10.0 \\

\midrule

\multirow{6}{*}{Diffusion policy }
&Action chunk size & 4 \\
&Input dimension & 50 \\
&Diffusion step embedding dimension & 256 \\
&Number of ResNet blocks & 3 \\
&Dropout rate & 0.1 \\
&LayerNorm & \cmark \\
\midrule

\multirow{6}{*}{Encoder}
&Encoder dimension & 256 \\
&Activation Function & ELU \\
&LayerNorm at each layer & \cmark \\
&SimNorm dimension & 5 \\
&Number of channels & 32 \\
&Latent dimension & 50 \\
\midrule

\multirow{6}{*}{MPC/ MPPI }
&Prediction horizon & 4 \\
&Execution horizon & 1 \\
&Number of iterations & 6 \\
&Number of samples & 512 \\
&Number of elites & 64 \\
&Inverse-temperature parameter $\tau$ & 0.01 \\
\midrule

\multirow{2}{*}{Reward and Latent dynamics}
&MLP dimension & 512 \\
&Activation Function & Mish \\
&LayerNorm at each layer & \cmark \\
\midrule

\multirow{4}{*}{Q, V and A functions }
&MLP dimension & 512 \\
&Activation Function & ELU, Tanh \\
&LayerNorm at each layer & \xmark \\
&Gradient reduction $\varrho$ & 0.8 \\
\midrule

\end{tabular}
\end{table}


\newpage

\section{Comparison of training curves between \method and baselines.}
\label{app:curves}

\begin{figure}[htbp]
    \centering
    \makebox[\textwidth][c]{\includegraphics[width=0.95\linewidth]{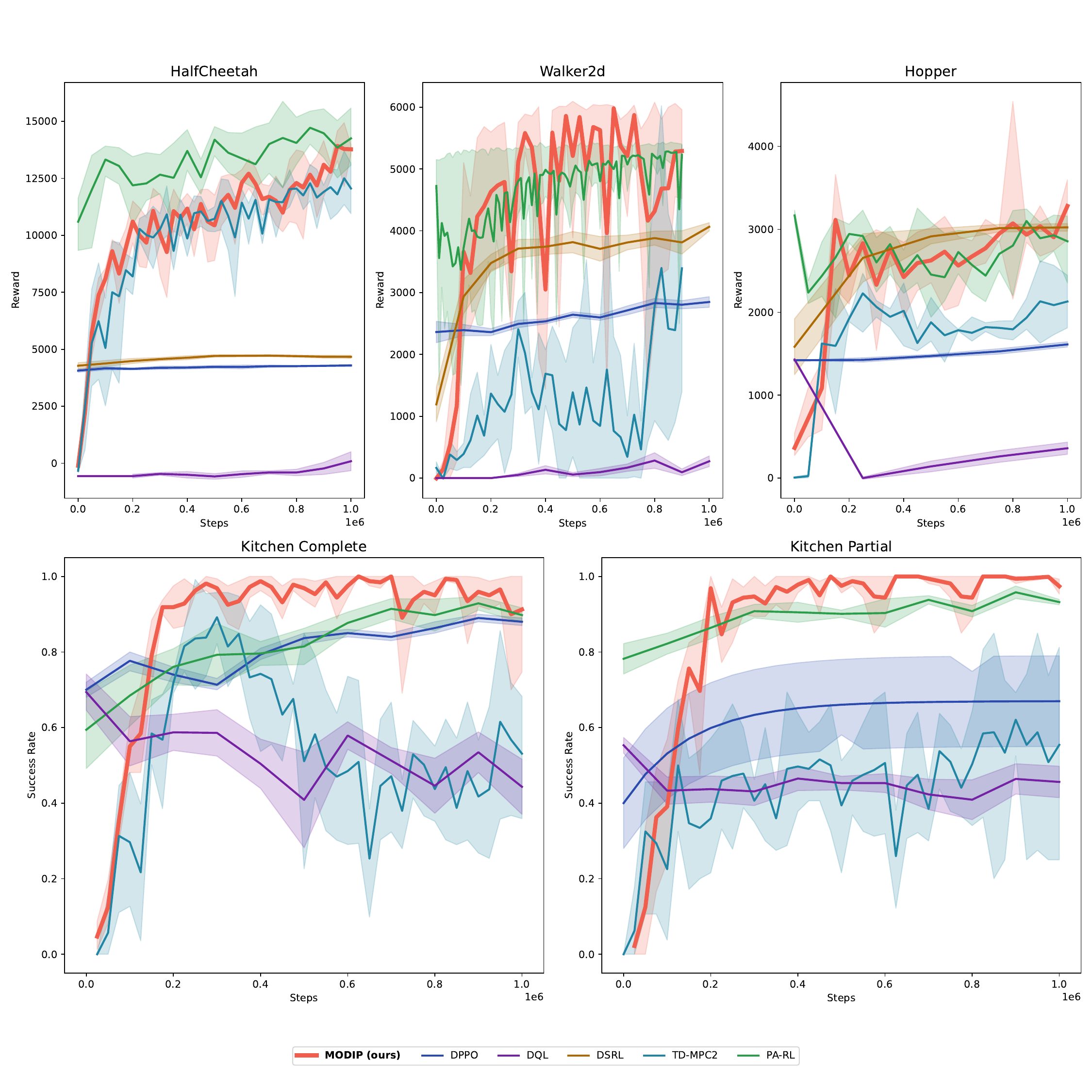}}
    \caption{Learning curve during fine-tuning, comparison of \method against baselines across MuJoCo locomotion tasks and Franka Kitchen manipulation environments.}
    \label{fig:mosaic}
\end{figure}

Figure~\ref{fig:mosaic} shows the online learning curves across MuJoCo and Kitchen tasks over a budget of $10^6$ environment steps, comparing \method against diffusion-policy fine-tuning methods and strong model-based baselines. Overall, \method improves rapidly during online fine-tuning and reaches strong final performance across environments. On Kitchen tasks, \method clearly outperforms the baselines in both sample efficiency and asymptotic success rate. On MuJoCo, \method is competitive with or better than the strongest baselines: it achieves strong performance on HalfCheetah, reaches the best final return on Walker2d, and is among the top methods on Hopper.

Interestingly, \method and PA-RL are the strongest methods on HalfCheetah, while chunk-based diffusion-policy RL baselines tend to plateau at lower returns. This suggests that locomotion tasks may benefit from frequent feedback and action correction: PA-RL optimizes single-step actions, whereas \method replans with MPC, allowing it to adapt actions more reactively than methods that execute longer diffusion-generated chunks.

Compared to TD-MPC2, \method consistently achieves stronger final performance in these learning curves. While TD-MPC2 relies on a standard Gaussian MLP policy prior, \method uses an expressive diffusion-policy prior inside MPC, which appears to provide more informative candidate action sequences and faster policy improvement.

\newpage

\end{document}